\def\year{2018}\relax
\documentclass[letterpaper]{article} 
\usepackage{aaai18}  
\usepackage{times}  
\usepackage{helvet}  
\usepackage{courier}  
\usepackage{url}  
\usepackage{graphicx}  
\usepackage{multirow} 
\usepackage{array}

\begin{document}
%
\title{Reading Scene Text with Attention Convolutional Sequence Modeling}
\author{Yunze Gao$^{1,2}$, Yingying Chen$^{1,2}$, Jinqiao Wang$^{1,2}$, Hanqing Lu$^{1,2}$\\
$^{1}$National Lab of Pattern Recognition, Institute of Automation, Chinese Academy of Sciences, Beijing, China\\
$^{2}$University of Chinese Academy of Sciences, Beijing, China\\
\{yunze.gao, yingying.chen, jqwang, luhq\}@nlpr.ia.ac.cn\\
}
\nocopyright
\maketitle
\begin{abstract}
Reading text in the wild is a challenging task in the field of computer vision. Existing approaches mainly adopted Connectionist Temporal Classification (CTC) or Attention models based on Recurrent Neural Network (RNN), which is computationally expensive and hard to train. In this paper, we present an end-to-end Attention Convolutional Network for scene text recognition. Firstly, instead of RNN, we adopt the stacked convolutional layers to effectively capture the contextual dependencies of the input sequence, which is characterized by lower computational complexity and easier parallel computation. Compared to the chain structure of recurrent networks, the Convolutional Neural Network (CNN) provides a natural way to capture long-term dependencies between elements, which is 9 times faster than Bidirectional Long Short-Term Memory (BLSTM). Furthermore, in order to enhance the representation of foreground text and suppress the background noise, we incorporate the residual attention modules into a small densely connected network to improve the discriminability of CNN features. We validate the performance of our approach on the standard benchmarks, including the Street View Text, IIIT5K and ICDAR datasets. As a result, state-of-the-art or highly-competitive performance and efficiency show the superiority of the proposed approach.
\end{abstract}

\section{Introduction}
Scene text recognition is to reading text in natural images, which has received considerable attention and plays an important role in a variety of computer vision tasks. Reading text in the wild can extract rich semantic information that is highly relevant to scene or object and therefore has been applied in plenty of real world applications, such as street sign reading in the driveless vehicle, automatic car license plate recognition, assistive technologies for the blind, robot navigation, scene understanding and image retrieval. However, suffering from various appearance, distortion, low resolution, blur and disturbance of background noise, text recognition in unconstrained environment is still a challenging problem.\par
\begin{figure}
\centering
\includegraphics[width=8.7cm]{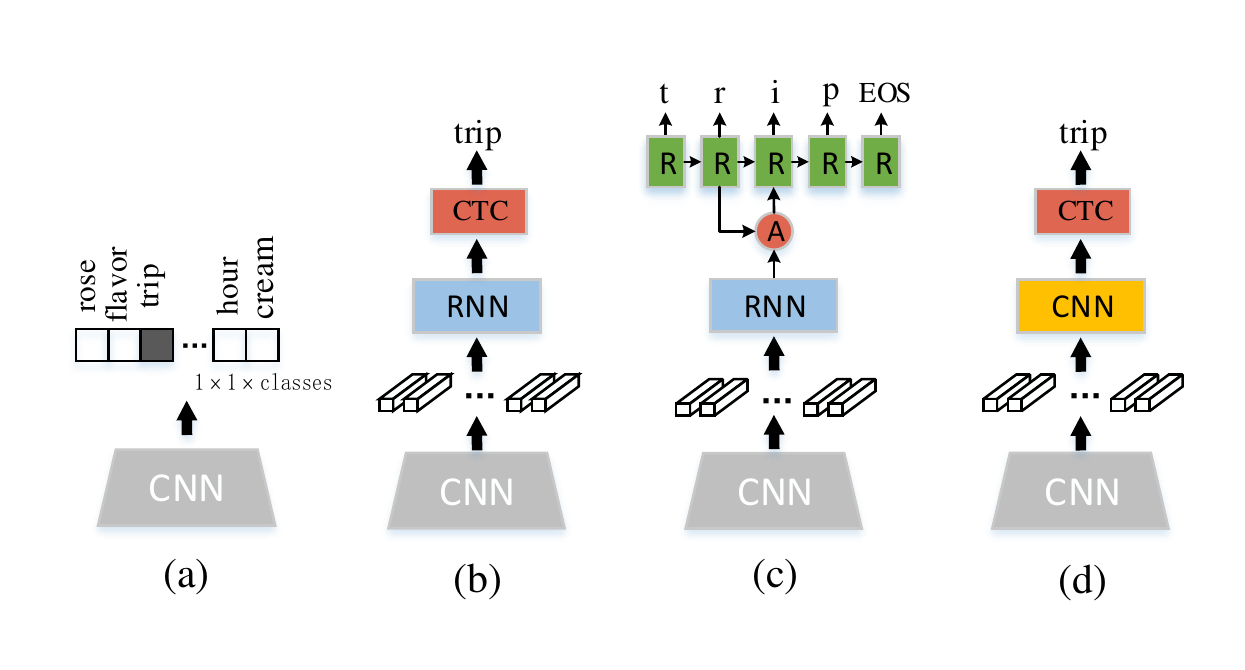}
\caption{Three typical architectures and our network for scene text recognition. (a) CNN + softmax. (b) RNN + CTC. (c) RNN + Attention. (d) CNN + CTC (our approach).}
\label{fig:picture001}
\end{figure}

Traditional methods \cite{wang2011end,bissacco2013photoocr,mishra2012top} recognized scene text by first detecting individual character and then recognizing each cropped character with convolutional neural network separately. A large amount of inter-character and intra-character confusion impact the performance of the entire recognition network greatly. Therefore, these approaches rely on an accurate character detector. Recently, some works adopted an end-to-end CNN framework for scene text recognition. \cite{jaderberg2016reading} formulated the scene text recognition as an image classification problem. As shown in Figure \ref{fig:picture001}(a), each class corresponds to one word in a pre-defined large lexicon composed of around 90k words. However, it is difficult to be generalized to other situations that has huge number of classes due to the oversize pre-refined dictionary and the requirement for large amount of training samples.\par

\begin{figure*}
\centering
\includegraphics[width=17.5cm]{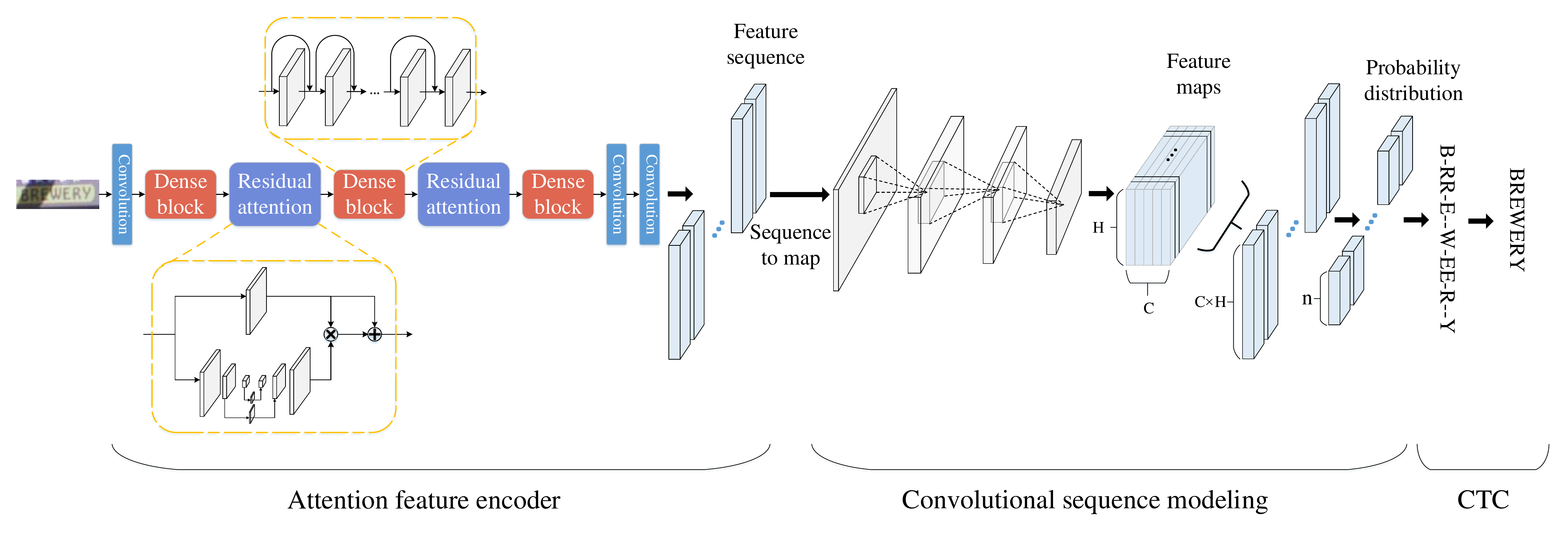}
\caption{Overview of the attention convolutional network. The attention feature encoder is a small densely connected network with residual attention for extracting feature sequence, in which each vector of the feature sequence corresponds to a local region of the input image. Then the elements of the feature sequence are put together to form a 2D map, which serves as the input of a convolutional neural network. Next, CNN captures the effective contextual information and learns the sequential dependencies, which is characterized by lower computational complexity and easier parallel computation. Subsequently, the output feature maps are restored to a sequence as the input of CTC to generate final label sequence. Specifically, ``n'' denotes the number of classes.}
\label{fig:picture002}
\end{figure*}

To address this dilemma, recent studies \cite{shi2016end,lee2016recursive,shi2016robust} regarded scene text recognition as a sequence recognition problem to directly generate label sequence, by means of recurrent neural network with CTC \cite{graves2006connectionist} or Attention schemes. As shown in Figure \ref{fig:picture001}(b), RNN is leveraged to model the sequential dependencies within the feature sequence produced by CNN and then CTC is to predict label sequence with arbitrary length. Afterwards, attention decoder \cite{shi2016robust,lee2016recursive} was proposed to weight the feature sequence and perform soft features selection, as illustrated in Figure \ref{fig:picture001}(c). \par

RNN is effective to learn the contextual information and capture the long-term dependencies. However, depending on the computation of the previous step, the recurrent connection is incapable of performing parallel operations. The process of sequence modeling is closely related to the length of input sequence due to the chain structure, so recurrent network is computation intensive and time consuming. Besides, sometimes RNN is difficult to train due to the problem of gradient vanishing/exploding \cite{bengio1994learning}. By contrast, CNN has the superiority of parallelism and lower computational complexity. Convolutional layers apply filters over the entire sequence simultaneously and allow parallel computation. Recent works have used CNN for sequence learning in machine translation \cite{gehring2017convolutional} and language modeling \cite{dauphin2016language}. In this paper, we present an end-to-end attention convolutional network for scene text recognition, which combines CNN with CTC to generate label sequence, without any recurrent connections, as shown in Figure \ref{fig:picture001}(d). Firstly, a sequence-to-map operation transforms the feature sequence into a 2D map as the input of CNN to process the sequence simultaneously. Then stacked convolutional layers can extract hierarchical contextual representation of the input sequence to model the long-term dependencies with a shorter path. At the same time, the length of dependencies can be controlled by the number of convolutional layers. This architecture is entirely convolutional, so it is easily for parallel processing and independent of the sequence length. \par

Besides, in order to enhance the representation of the text to be recognized and suppress the background noise, we incorporate a residual attention mechanism into a small densely connected network to extract the discriminative attention-aware features. We evaluate our approach on challenging benchmarks including the Street View Text, IIIT-5K and ICDAR datasets. It is observed that the proposed method can improve the efficiency with the state-of-the-art performance. The mainly contributions are summarized as follows:\par
(1) A novel end-to-end attention convolutional network is proposed for scene text recognition, which is entirely convolutional and performs well with both high accuracy and efficiency.\par
(2) Instead of RNN, we adopt the stacked convolutional layers to effectively capture the contextual dependencies of the input sequence, which is characterized by lower computational complexity and easier parallel computation.\par
(3) The residual attention modules in the small densely connected network could effectively suppress the response of background noise while enhancing the representation of foreground text.

\section{Related Work}

Traditional methods of scene text recognition first performed detection to generate multiple candidates of character locations, then applied a character classifier for recognition. \cite{wang2011end} used Random Ferns and HOG features to detect characters and then found an optimal configuration of a particular word via a pictorial structure. \cite{mishra2012top} detected character candidates using sliding windows and integrated both bottom-up and top-down cues in a unified Conditional Random Field (CRF) model. \cite{shi2013scene} constructed a part-based tree-structured model to recognize characters in cropped images. \cite{yao2014strokelets} proposed an alternative way for character representation, denoted as Strokelets, which was a combination of multi-scale mid-lever features.\par

Afterwards, the explorations of scene text recognition focus on the mapping from the entire image to word string directly. \cite{almazan2014word} embedded word images and word labels into a common Euclidean space and the embedding vectors were used to match images and labels. \cite{jaderberg2014deep} constructed two CNNs to classify character at each position in the word and detect the N-grams contained within the word separately, following a CRF model to combile their representations. Recently, there are increasing researches on treating scene text recognition as a sequence recognition problem. \cite{shi2016end} proposed Convolutional Recurrent Neural Network (CRNN) which combined convolutional network and recurrent network to model the spatial dependencies. In \cite{lee2016recursive}, a recurrent network with attention mechanism was constructed to decode feature sequence and predict labels recurrently. \cite{shi2016robust} adopted a convolutional-recurrent structure in the encoder to learn the sequential dynamics. \par

Since the lower computational complexity and greater parallelism, CNN is a more efficient structure to capture the sequential contextual information. Some attempts for applying CNN to sequence modeling have been made to replace RNN. \cite{dauphin2016language} introduced a new neural language model that replaced the recurrent connections typically used in RNN with gated temporal convolutions. \cite{strubell2017fast} proposed the Iterated Dilated Convolutional Neural Networks (ID-CNNs), which was a faster alternative to recurrent network for obtaining per token vector representations in Natural Language Processing (NLP). \cite{gehring2017convolutional} proposed an architecture based entirely on convolutional neural network in machine translation. Although CNN has shown the superiority of parallelism and efficiency, as far as we know, there is no research on using CNN to perform sequence generation in the field of scene text recognition. Our method incorporates CNN and CTC into a unified framework without any recurrent connections, which improves the efficiency while achieving good performance. Furthermore, the proposed algorithm is not limited by a pre-refined dictionary and is available in both lexicon-free and lexicon-based setting. Without individual character detection, the proposed network can be trained end-to-end with the word level annotations and can effectively deal with words with arbitrary length.

\section{The Proposed Approach}

The overview of the attention convolutional network for scene text recognition is illustrated in Figure \ref{fig:picture002}, which is composed of an attention feature encoder, a convolutional sequence modeling module and a CTC module. The attention feature encoder is a small densely connected network with residual attention for extracting feature sequence. Specifically, The encoder takes the cropped word image as input and extracts robust feature representation for the entire image, with a residual attention mechanism to suppress background noise. Then the feature maps produced by the dense attention network are converted into a feature sequence. Each element in the feature sequence, which corresponds to a local region of the word image, can be viewed as the feature representation of the region. In the convolutional sequence modeling module, in order to process the sequence simultaneously, a sequence-to-map operation transforms the feature sequence into a 2D map to serve as the input of the convolutional layers. Then stacked convolutional layers are leveraged to capture the effective contextual information and learn the sequential dependencies. Next, the same as the map-to-sequence operation in the encoder, the output feature maps are restored to a sequence as the input of CTC. Finally, CTC is used to perform sequence generation with arbitrary length. In the following sections, we will describe the three components in detail.

\subsection{Attention Feature Encoder}

To improve the discriminability of CNN features, we design a dense attention encoder network to extract the attention-aware representation. In dense blocks, dense connectivity can improve the information flow between layers, and meanwhile, equipped with a residual attention mechanism, the disturbance of background noise can be effectively reduced. Furthermore, we convert the feature maps into a sequential representation of the input image. Specifically, in a left-to-right order, we take out the same columns of feature maps and concatenate them into a vector, which corresponds to a local rectangular region of the word image.

\subsubsection{Dense Connectivity}

Taking advantage of the power of Densenet \cite{huang2016densely}, we utilize the dense connectivity to improve the flow of information and gradient propagation in the encoder network. There exists direct connections between all layers in the dense block. Therefore, each layer can get the information from all preceding layers and transmit its message to all subsequent layers. Additionally, instead of gradient back propagation from layer to layer, each layer can obtain the deep supervision, which eases the learning. The concatenation of feature maps produced by preceding layers severs as the input of the later layers.

\subsubsection{Residual Attention}
Attention mechanism plays an important guiding role in the process of feature learning, which aims to selectively focus on the salient regions of the objects and enhance the representation of relevant parts. For text in natural images, there often exists some disturbances, including shadow, irrelevant symbols and background texture. The scene text with various appearance is often confused by these factors. Therefore, we introduce a residual attention mechanism as \cite{wang2017residual} to enhance the representation of foreground text and suppress background noise. \par

The residual attention module serves as the transition between dense blocks, which is divided into two branches. Specifically, the feature branch performs the feedforward process and the attention branch generates the soft attention weights. The attention branch is designed with bottom-up top-down structure so that the high-level semantic information can be captured to guide the discriminative feature selection. Max pooling followed by the convolutional layer are stacked several times to increase the receptive field rapidly and collect the global information. Then a symmetrical architecture with bilinear interpolation for upsampling is applied to recover the resolution. Afterwards, the attention maps as soft weights are added on corresponding feature maps in each position. Considering the attention weights range from zero to one, the element-wise product between feature maps and attention maps may cause severe degradation of useful information. Therefore, the residual attention learning is utilized to address this issue. Similar to the connection in Resnet \cite{he2016deep}, the output of residual attention module is:\par
\begin{equation}
T = (1+A) \times F
\end{equation}
where $F$ and $A$ denote the output of the feature branch and the attention branch, respectively. In this way, the background noise can be suppressed efficiently while enhancing the discriminability of original features. Furthermore, different attention modules generate the attention maps adapted to the corresponding features. The low-level attention module mainly concentrates on the appearance including edge, color and texture, while the high level attention module contains more semantic information. With the residual attention mechanism, the feature encoder benefits from the noise suppression to obtain a more discriminative representation.

\subsection{Convolutional Sequence Modeling}

As the dominant approach of sequence to sequence learning, RNN has been successfully applied in many computer vision tasks, including speech recognition \cite{hannun2014deep,miao2013deep}, language modeling \cite{jozefowicz2016exploring} and machine translation \cite{luong2014addressing}. However, depending on the computation of the previous step, the recurrent connection is incapable of performing parallel operations. Besides, sometimes RNN is difficult to train due to the problem of gradient vanishing/exploding. Therefore, in this paper, we adopt CNN to capture the sequential dependencies with both directions for scene text recognition, which is a faster alternative of recurrent network. Given the feature sequence produced by encoder, which is denoted as $\textbf{f} = (f_{1}, f_{2}, \cdots, f_{w})$. In order to get the contextual information $\textbf{c} = (c_{1}, c_{2}, \cdots, c_{w})$, the recurrent neural network $R$ generates the contextual representation through the recurrent connection $c_{i} = R(c_{i-1}, f_{i})$, which is a chain structure and is unable to perform parallel computation. \par

Our approach models the sequential dependencies with entirely convolutional operation. Firstly, the elements of the feature sequence are put together to form a 2D map, where each column is associated with a local region of original word image from left to right. Then we convolve the input using a filter with width $k$, resulting contextual information over k elements of the input sequence. Stacked convolutional layers generate the hierarchical representation to enlarge the size of receptive field rapidly. For example, 4 convolutional layers with kernel size $k=5$ can capture the information of 17 input elements. In this way, we can easily control the range of spatial dependencies to be modeled through the number of convolutional layers. When there are sufficient layers, the high-level features can obtain the requiring contextual information. Besides, the convolution operation does not depend on the state of the previous step and is irrelevant to the length of input sequence. Therefore, the computation over the entire sequence can be simultaneously processed, which could greatly accelerate the process of sequence modeling. Furthermore, the convolutional network consumes less memory space and running time due to fewer parameters and lower computation complexity.

In the process of sequence modeling, we ensure that the length of sequence remains unchanged, by means of convolution with zero padding. Afterwards, in order to get the sequential representation to serve as the input of CTC, we restore the output feature maps to a sequence again by the same map-to-sequence operation in the feature encoder. Assuming the feature maps produced by CNN have the dimension of $C \times H \times W$, where $C$, $H$ and $W$ denote the channels, height and width, respectively. Specifically, we crop the each channel of feature maps by column and then concatenate the same columns of all channels into a vector, which has a dimension of $C \times H$. Therefore, we can obtain a sequence with $W$ vectors, which is the contextual information $\textbf{c} = (c_{1}, c_{2}, \cdots, c_{w})$. Finally, for the generated sequence, we obtain the probability distribution over the label space for per-frame in the sequence via a linear layer:
\begin{equation}
y_{t}=softmax(W*c_{t}+b), t=1, \cdots, w
\end{equation}
where $W$ and $b$ denote the weight matrix and bias separately.

\subsection{Connectionist Temporal Classification}
To generate the final label sequence for the input image, the output probability distribution sequence $\textbf{y} = (y_{1}, y_{2}, \cdots, y_{w})$ is interpreted as conditional probabilities over possible label sequences, using the Connectionist Temporal Classification (CTC) proposed by \cite{graves2006connectionist}. Defining $L$ as the set of 36 classes including all English alphanumeric characters, we get the final label space $L^{'}=L \cup { \left\{ blank \right\} }$, in which the extra $blank$ denotes the class for observing no character. Given the probability distribution, the conditional probability of the sequence $\pi$ is
\begin{equation}
p( \pi | \textbf{y}) = \prod^{w}_{t=1} y^{t}_{\pi_{t}}
\end{equation}
where $y^{t}_{\pi_{t}}$ denotes the probability of emtting label $\pi_{t}$ at step $t$. Then a many-to-one mapping $\mathcal{B}$ is defined to transform the sequence $\pi$ to a shorter sequence as the final prediction. The operation $\mathcal{B}$ is to merge the repeated continuous labels to a single one and then remove the $blank$ labels. For example, $\mathcal{B}$ maps the sequence ``- -aa-b- -c-dd'' to ``abcd'', where the `-' represents the $blank$. Different sequences $\pi$ may be mapped to the same result, thus the probability of the final output sequence is formulated as the sum of the conditional probabilities of all $\pi$ corresponding to it:

\begin{equation}
p( l |\textbf{y}) = \sum_{\pi \in \mathcal{B}^{-1}(l)}p(\pi | \textbf{y})
\label{equ4}
\end{equation}

In genaral, for a given sequence, there are a large number of mapping paths. Thus the computation of the sum in Equation \ref{equ4} is time consuming. To remedy this issue, the forward-backward algorithm based on dynamic programming is used to calculate the conditional probability and the error differentials in an efficient way.\par
Given the training set $\mathcal{D}=\left\{I_{i},l_{i} \right\}$, where $I_{i}$ and $l_{i}$ represent the word image and the corresponding ground truth label, respectively. The objective function is formulated as the sum of the negative log likelihood of the probabilities for target labels:

\begin{equation}
\mathcal{O} = - \sum_{(I_{i},l_{i}) \in \mathcal{D}} \log p(l_{i}|\textbf{y}_{i})
\end{equation}

Minimizing the objective function equals to maximize the probability of producing the target labels, which can be solved by dynamic programming. By means of CTC, we can handle sequences with arbitrary length, requiring no pre-segmented training data.

\subsubsection{Lexicon-free Recognition}

Our network can be used in the lexicon-free setting, i.e., the predicted words are not constrained with a pre-defined dictionary. During inference, the sequence with the highest conditional probability severs as the final output. However, there is no a general and tractable approach to find the optimal solution. So the approach of the best path decoding \cite{graves2006connectionist} is applied to get the output label sequence approximately, assuming that the most probable path corresponds to the most probable label:

\begin{equation}
l_{p}=\mathcal{B}(\mathop{argmax} \limits_{\pi}p(\pi|\textbf{y}))
\end{equation}

Specifically, the most probable path $\pi$ is produced by emitting the label with the highest probability at each step.

\subsubsection{Lexicon-based Recognition}

In the constrained condition, the prediction is generated by selecting the word with the highest conditional probability in the pre-defined lexicon. However, the probabilities need to be computed for all words in the dictionary, which is time consuming and infeasible. So we adopt an approximate method by comparing the edit distance between the predicted sequence in the lexicon-free setting and words in the lexicon, then choosing the word with the smallest edit distance as the output label.

\section{Experiment}
In this section, we will give the implementation details and evaluate the performance of the proposed approach.

\subsection{Dataset}
Several public datasets are used for the evaluation, including Street View Text, IIIT5K, ICDAR 2003 and ICDAR 2013.
\begin{itemize}
\item \textbf{Street View Text} \cite{wang2011end} contains 647 word images which are cropped from 249 street-view images collected from Google Street View. For each image, there is a 50 words lexicon defined by \cite{wang2011end}, denoted as SVT-50.
\item \textbf{IIIT5K} \cite{mishra2012scene} contains 3000 cropped word images collected from the Internet. Each image has a 50 words lexicon and a 1000 words lexicon, denoted as IIIT5k-50 and IIIT5k-1k separately.
\item \textbf{ICDAR 2003} \cite{lucas2005icdar} contains 251 scene images and 860 cropped word images. The 50 words lexicon for each image is defined by \cite{wang2011end}, which is denoted as IC03-50. And a full lexicon is composed of all the words that appear in the test set, denoted as IC03-Full.
\item \textbf{ICDAR 2013} \cite{karatzas2013icdar} contains 1015 cropped word images without any lexicon, which derives from the ICDAR 2003.
\end{itemize}
For training data, our model is trained purely on the synthetic dataset released by \cite{jaderberg2014synthetic}, without any extra data for fine-tuning. The synthetic dataset contains around 8 millions images, which are generated by an synthetic engine. Following the evaluation protocol in \cite{wang2011end}, we perform recognition on word images that contain only alphanumeric characters and at least three characters.

\begin{table}
  \centering
  \caption{Network architecture of the attention convolutional network. The growth rate is 18 in the dense blocks. Stride 2$\times$1 represents the filter with height 2 and width 1.}
  \begin{tabular}{c|c|c}
    \hline
    Module & Layer & Configurations\\
    \hline
    \multirow{12}* {Encoder}
    & Convolution & 3$\times$3, 36, stride 1$\times$1\\
    \cline{2-3}
    & Dense Block & [3$\times$3, stride 1$\times$1] $\times$ 4\\
    \cline{2-3}
    & Attention Module & Attention 1\\
    \cline{2-3}
    & Average Pooling & 2$\times$2, stride 2$\times$2 \\
    \cline{2-3}
    & Dense Block  & [3$\times$3, stride 1$\times$1] $\times$ 4\\
    \cline{2-3}
    & Attention Module & Attention 1\\
    \cline{2-3}
    & Average Pooling & 2$\times$2, stride 2$\times$2 \\
    \cline{2-3}
    & Dense Block  & [3$\times$3, stride 1$\times$1] $\times$ 4 \\
    \cline{2-3}
    & Convolution & 3$\times$3, 512, stride 1$\times$1  \\
    \cline{2-3}
    & Average Pooling & 2$\times$2, stride 2$\times$1 \\
    \cline{2-3}
    & Convolution & 3$\times$3, 512, stride 1$\times$1  \\
    \hline
    \multirow{4}* {CNN}
    & Convolution & 3$\times$3, 1, stride 2$\times$1  \\
    \cline{2-3}
    & Convolution & 3$\times$3, 1, stride 2$\times$1 \\
    \cline{2-3}
    & Convolution & 3$\times$3, 1, stride 2$\times$1  \\
    \cline{2-3}
    & Convolution & 3$\times$3, 1, stride 2$\times$1  \\
    \hline
    CTC & CTC & -\\
    \hline
\end{tabular}
\label{tab:001}
\end{table}

\begin{table*}
  \centering
  \caption{Scene text recognition accuracies on the benchmark datasets. ``50'', ``1000'' and ``Full'' represent the size of lexicon used for constrained recognition. ``$\ast$''\cite{jaderberg2016reading} is not lexicon-free strictly, due to the output sequence is constrained to a 90k dictionary.}

\begin{tabular}{p{4.39cm}<{\centering}|p{1.15cm}<{\centering}|p{0.6cm}<{\centering}|p{1.4cm}<{\centering}|p{1.4cm}<{\centering}|p{0.8cm}<{\centering}|p{1.2cm}<{\centering}|p{1.45cm}<{\centering}|p{0.6cm}<{\centering}|p{0.55cm}<{\centering}}
  \hline
  Methods & SVT-50 & SVT & IIIT5k-50 & IIIT5k-1k & IIIT5k & IC03-50 & IC03-Full & IC03 & IC13 \\
  \hline
  ABBYY\shortcite{wang2011end} & 35.0 & - & 24.3 & - & - & 56.0 & 55.0 & - & - \\
  Wang \emph{et al}.\shortcite{wang2011end} & 57.0 & - & - & - & - & 76.0 & 62.0 & - & - \\
  Mishra \emph{et al}.\shortcite{mishra2012scene} & 73.2 & - & 64.1 & 57.5 & - & 81.8 & 67.8 & - & - \\
  Wang \emph{et al}.\shortcite{wang2012end} & 70.0 & - & - & - & - & 90.0 & 84.0 & - & - \\
  Goel \emph{et al}.\shortcite{goel2013whole} & 77.3 & - & - & - & - & 89.7 & - & - & - \\
  Bissacco \emph{et al}.\shortcite{bissacco2013photoocr} & 90.4 & 78.0 & - & - & - & - & - & - & 87.6 \\
  Alsharif and Pineau \shortcite{alsharif2013end} & 74.3 & - & - & - & - & 93.1 & 88.6 & - & - \\
  Almaz\'an \emph{et al}.\shortcite{almazan2014word} & 89.2 & - & 91.2 & 82.1 & - & - & - & - & - \\
  Yao \emph{et al}.\shortcite{yao2014strokelets} & 75.9 & - & 80.2 & 69.3 & - & 88.5 & 80.3 & - & - \\
  Rodriguez-Serrano \emph{et al}.\shortcite{rodriguez2015label} & 70.0 & - & 76.1 & 57.4 & - & - & - & - & - \\
  Jaderberg \emph{et al}.\shortcite{jaderberg2014spotting} & 86.1 & - & - & - & - & 96.2 & 91.5 & - & - \\
  Su and Lu \shortcite{su2014accurate} & 83.0 & - & - & - & - & 92.0 & 82.0 & - & - \\
  Gordo \shortcite{gordo2015supervised} & 91.8 & - & 93.3 & 86.6 & - & - & - & - & - \\
  $\ast$Jaderberg \emph{et al}.\shortcite{jaderberg2016reading} & 95.4 & 80.7 & 97.1 & 92.7 & - & \textbf{98.7} & \textbf{98.6} & \textbf{93.1} & \textbf{90.8} \\
  Jaderberg \emph{et al}.\shortcite{jaderberg2014deep} & 93.2 & 71.7 & 95.5 & 89.6 & - & 97.8 & 97.0 & 89.6 & 81.8 \\
  Shi \emph{et al}.\shortcite{shi2016end} & \textbf{97.5} & \textbf{82.7} & 97.8 & 95.0 & 81.2 & \textbf{98.7} & 98.0 & 91.9 & 89.6 \\
  Shi \emph{et al}.\shortcite{shi2016robust} & 95.5 & 81.9 & 96.2 & 93.8 & \textbf{81.9} & 98.3 & 96.2 & 90.1 & 88.6 \\
  Lee \emph{et al}.\shortcite{lee2016recursive} & 96.3 & 80.7 & 96.8 & 94.4 & 78.4 & 97.9 & 97.0 & 88.7 & 90.0 \\
  Ghosh \emph{et al}.\shortcite{ghosh2017visual} & 95.2 & 80.4 & - & - & - & 95.7 & 94.1 & 92.6 & - \\
  \hline
  Ours & 97.4 & \textbf{82.7} & \textbf{99.1} & \textbf{97.9} & 81.8 & \textbf{98.7} & 96.7 & 89.2 & 88.0 \\
  \hline
\end{tabular}
\label{tab:004}
\end{table*}

\subsection{Implementation Details}

The network architecture of our approach is shown in Table \ref{tab:001}. The attention feature encoder is composed of three dense blocks and two residual attention modules. For each dense block, there are four convolutional layers and the growth rate is 18. For residual attention, the feature branch has one convolutional layer and the attention branch has the bottom-up top-down structure. In detail, 3,2 max pooling are used in the two residual attention modules, respectively. And a sigmoid function is applied to normalize the weights in attention maps. Besides, the skip connection is added to fuse information with different scales. Additionally, the CNN for sequence modeling contains 4 convolutional layers, so that the per-frame representation of the output sequence can cover the contextual information over 9 input elements. All the convolution are performed with zero padding, ReLU activation function and batch normalization \cite{ioffe2015batch}. \par

In the process of training and testing, the word images are resized to $32 \times 100$ with gray scale. We adopt the msra \cite{he2015delving} as the weight initialization method and train our network using Adam \cite{kingma2014adam} with a mini-batch size of 64. Moreover, the gradient clipping is used at the magnitude of 5. The proposed network is implemented with Tensorflow \cite{abadi2016tensorflow}.

\subsection{Ablation Study}

In this section, we evaluate the contributions of the two main components in the proposed approach, including residual attention modules for suppressing noise and convolutional network for sequence modeling.

\subsubsection{Residual Attention Module}
\begin{figure}
\centering
\includegraphics[width=8.5cm]{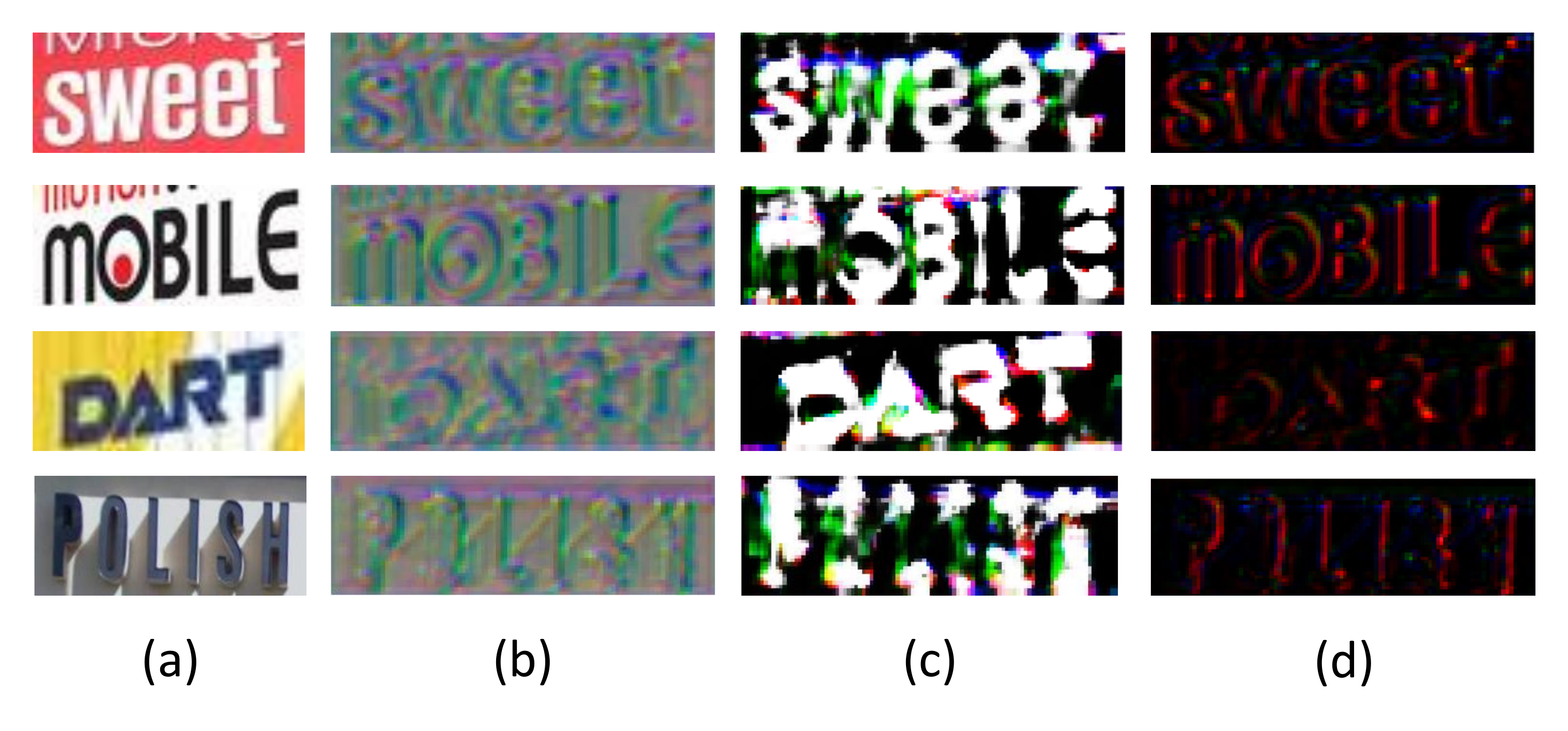}
\caption{Visualization of the feature maps and attention maps. (a) Original images. (b) Feature maps before attention. (c) Attention maps. (d) Feature maps after attention.}
\label{fig:picture003}
\end{figure}

To validate the effectiveness of residual attention modules, we explore the performance of densely connected network with BLSTM and CNN while applying residual attention mechanism. As shown in Table \ref{tab:002}, the networks with residual attention modules consistently outperform the networks without attention, which proves the effectiveness of the method. Besides, we observe that the improvements brought by residual attention modules on ICDAR datasets are not as significant as that on SVT and IIIT5k. The possible reason is that the proportion of images with background noise in SVT and IIIT5k is higher than that in ICDAR datasets. Furthermore, we visualize the attention maps of some examples in Figure \ref{fig:picture003}. As shown in Figure \ref{fig:picture003}, the attention maps focus on the foreground text to be recognized and effectively reduce the response of background noise including shadow, irrelevant symbols, and background texture.

\begin{table}
  \centering
  \caption{Lexicon-free scene text recognition accuracies on standard benchmarks. The growth rate is 18 for all networks.}

\begin{tabular}{p{3.5cm}<{\centering} | p{0.6cm}<{\centering} | p{0.75cm}<{\centering} | p{0.6cm}<{\centering} | p{0.55cm}<{\centering}} 
  \hline
  Method&SVT&IIIT5k&IC03&IC13 \\
  \hline
  Dense+BLSTM&81.9&80.4&90.7& 89.5 \\
  Dense+BLSTM+attention&83.0&83.0&90.8&88.7 \\
  \hline
  Dense+CNN&81.2  & 79.6 & 88.3 & 88.6 \\
  Dense+CNN+attention&82.7&81.8&89.2&88.0 \\
  \hline
\end{tabular}
\label{tab:002}
\end{table}

\subsubsection{Convolutional vs. Recurrent Model}
To compare the performance of recurrent network and convolutional network, we replace the convolutional layers for sequence modeling with 2 layers BLSTM which has 256 units per layer as the baseline. As shown in Table \ref{tab:003}, compared with BLSTM, the convolutional neural network requires significantly fewer parameters and less time to achieve comparable performance. Furthermore, the speed of sequence modeling with CNN can run 9 times faster than BLSTM and the parameters are greatly reduced. Therefore, the proposed CNN model significantly improves the efficiency while maintaining good performance. The convolutional sequence modeling module also can be used in other networks to capture contextual dependencies and achieve high efficiency. \par

\subsection{Comparisons with State-of-the-art Methods}

We evaluate the attention convolutional network on the above four public datasets and compare it with state-of-the-art algorithms in Table \ref{tab:004}. Most existing approaches were incapable of performing recognition without a dictionary and did not report the results in the unconstrained condition. By contrast, our method is available in both lexicon-free and lexicon-based setting.

For lexicon-free recognition, our network achieves the state-of-the-art or highly competitive performance. Specifically, we obtain the best result on SVT and the second best result on IIIT5k. It is worthy noting that the model in \cite{jaderberg2016reading} only can deal with the words in its 90k dictionary, which is not lexicon-free recognition strictly. Different form \cite{jaderberg2016reading}, our approach is able to recognize random word strings, which is not restricted by a fixed dictionary. Besides, we obtain the better or comparable results compared with \cite{shi2016robust}, while \cite{shi2016robust} used Spatial Transformer Network (STN) to rectify the irregular text. We do not perform any special operation aimed at irregular text, which shows the tolerance of our network to spatial distortions of scene text. Furthermore, most methods used RNN to capture contextual dependencies, while we adopt convolutional sequence modeling to achieve higher efficiency and competitive results. \par

For lexicon-based recognition, our method consistently outperforms other approaches on several benchmarks. Especially, we outperform prior state-of-the-art \cite{shi2016end} by margin of nearly 3 percentages on IIIT5k with 1000 words lexicon. The significant improvement validates the effectiveness of our method. Moreover, it is observed that IIIT5k contains plenty of images suffering from background noise, which proves the superiority of our method in suppressing noise. Besides, we only behind \cite{jaderberg2016reading} on ICDAR03 with the full lexicon. However, \cite{jaderberg2016reading} benefits from the pre-defined large dictionary as mentioned before. Therefore, our results are still competitive compared with the state-of-the-arts.

\begin{table}
  \centering
  \caption{Comparison between CNN and BLSTM. The SM time represents the time for sequence modeling (SM) with CNN or BLSTM.}

\begin{tabular}{c|c|c|c|c}
  \hline
  Model & SVT & Params & Total time & SM time \\
  \hline
  BLSTM-2L & 83.0 & 141.5M & 45.3ms & 31.7ms\\
  CNN-4L & 82.7 & 75.4M & 19.6ms & 3.5ms\\
  \hline
\end{tabular}
\label{tab:003}
\end{table}

\section{Conclusion}

In this paper, we propose a novel attention convolutional network for scene text recognition, which contains a densely connected network with residual attention modules for extracting features, stacked convolutional layers for sequence modeling and CTC for sequence generation. Instead of RNN, we introduce a convolutional neural network to capture the contextual information and model long-term dependencies with fewer parameters, which is 9 times faster than BLSTM. In addition, the usage of residual attention mechanism significantly improves the performance and suppresses background noise. Finally, the proposed network can be trained end-to-end with the word level annotations. And our method has capability of handling with word sequences with arbitrary length in both lexicon-free and lexicon-based setting. The extensive experimental results on the challenging datasets demonstrate the superiority of our approach compared with the state-of-the-art methods.

\bibliographystyle{aaai}
\bibliography{sample_bibliography}

\end{document}